\documentclass{article}
\usepackage{ijcai21}

\usepackage{times}
\usepackage{soul}
\usepackage{url}
\usepackage[utf8]{inputenc}
\usepackage[small]{caption}
\usepackage{graphicx}
\usepackage{amsmath}
\usepackage{amsthm}
\usepackage{booktabs}
\usepackage{algorithm}
\usepackage{algorithmic}
\usepackage{hyperref}
\usepackage{amsmath}
\usepackage{amsthm}
\newtheorem{mypro}{Proposition}
\newtheorem{lemma}{Lemma}
\usepackage{subfigure}
\usepackage[switch]{lineno}
\usepackage{multirow}
\usepackage{color}

\title{Regularizing Variational Autoencoder with Diversity and Uncertainty Awareness}

\author{
Dazhong Shen$^{1,2}$ \and
Chuan Qin$^{2}$\and
Chao Wang$^{1,2}$\and
Hengshu Zhu$^{2,*}$ \and
Enhong Chen$^1$ \and
Hui Xiong$^{3,}$\footnote{This work was done when Dazhong Shen was an intern at Talent Intelligent Center, Baidu Inc.  Hui Xiong and Hengshu Zhu are the corresponding authors.}
\affiliations
$^1$ School of Computer Science and Technology, University of Science and Technology of China\\
$^2$Baidu Talent Intelligence Center\\
$^3$Rutgers, The State University of New Jersey\\
\emails
sdz@mail.ustc.edu.cn,
chuanqin0426@gmail.com,
wdyx2012@mail.ustc.edu.cn,
zhuhengshu@baidu.com,
cheneh@ustc.edu.cn,
hxiong@rutgers.edu
}

\begin{document}

\maketitle

\begin{abstract}
As one of the most popular generative models, Variational Autoencoder (VAE) approximates the posterior of latent variables based on amortized variational inference. However, when the decoder network is sufficiently expressive, VAE may lead to \textit{posterior collapse}; that is, uninformative latent representations may be learned.  
To this end, in this paper, we propose an alternative model, DU-VAE, for learning a more \textit{\textbf{D}iverse} and less \textit{\textbf{U}ncertain} latent space, and thus the representation can be learned in a meaningful and compact manner. Specifically, we first theoretically demonstrate that it will result in better latent space with high diversity and low uncertainty awareness by controlling the distribution of posterior's parameters across the whole data accordingly. Then, without the introduction of new loss terms or modifying training strategies, we propose to exploit Dropout on the variances and Batch-Normalization on the means simultaneously to regularize their distributions implicitly. Furthermore, to evaluate the generalization effect, 
we also exploit DU-VAE for inverse autoregressive flow based-VAE (VAE-IAF) empirically. Finally, extensive experiments on three benchmark datasets clearly show that our approach can outperform state-of-the-art baselines on both likelihood estimation and underlying classification tasks.
\end{abstract}

\section{Introduction}
Recent years have witnessed the great success of Variational Autoencoder (VAE)~\cite{kingma2013auto} as a generative model for representation learning, which has been widely exploited in various challenging domains, such as natural language modeling and image processing~\cite{bowman2015generating,pu2016variational}. Indeed, VAE models the generative process of observed data by defining a joint distribution with latent space, and approximates the posterior of latent variables based on the amortized variational inference. While the use of VAE has been well-recognized,  it may lead to uninformative latent representations, particularly when  the expressive and powerful decoder networks are employed, such as LSTMs~\cite{hochreiter1997long} on text or PixelCNN~\cite{van2016conditional} on images. This is widely known as the \emph{posterior collapse} phenomenon~\cite{Zhao2019InfoVAEBL}. In other words, the model may fail to diversify the posteriors of different data by simply using the single posterior distribution component to model all data instances. Also, the traditional VAE  model usually produces the redundant information of representation due to the lack of guidance to characterize posterior space~\cite{bowman2015large,Chen2017VariationalLA}. Therefore, the learned representation of VAE often results in an unsatisfied performance for  downstream tasks, such as classification, even if it can approximate the marginal likelihood of observed data very well.
 
In the literature, tremendous efforts have been made for improving the representation learning of VAE and alleviating the problem of posterior collapse.
One thread of these works is to attribute the posterior collapse to optimization challenges of VAEs and design various strategies, including
KL annealing~\cite{bowman2015large,fu2019cyclical}, Free-Bits(FB)~\cite{kingma2016improved}, aggressive training~\cite{he2018lagging}, encoder network pretraining and decoder network weakening~\cite{yang2017improved}.
Among them, BN-VAE~\cite{Zhu2020ABN} applies the Batch-Normalization (BN)~\cite{ioffe2015batch} to ensure one positive lower bound of the KL term.
However, the theoretical basis of the effectiveness of BN on latent space learning is not yet understood, and more possible explanations based on the geometry analysis of latent space are needed. 
Other studies attempt to modify the objective carefully to direct the latent space learning~\cite{Makhzani2015AdversarialA,zheng2019understanding}. One feasible direction is to add additional Mutual Information (MI) based term to enhance the relation between data and latent space. However, due to the intractability, additional designs are always required for approximating MI-based objectives~\cite{fang2019implicit,Zhao2019InfoVAEBL}. Recently, Mutual Posterior-Divergence (MPD)~\cite{ma2018mae} is introduced to measure the diversity of the latent space, which is analytic and has one similar goal with MI. However, the scales of MPD and original objective are unbalanced, which requires deliberate normalization.


In this paper, to improve the representation learning performances of VAE, we propose a novel generative model, DU-VAE, for learning a more \textit{\textbf{D}iverse} and less \textit{\textbf{U}ncertain} latent space, and thus ensures the representation can be learned in a meaningful and compact manner. 
To be specific, we first analyze the expected latent space theoretically from two geometry properties, diversity and uncertainty, based on the MPD and Conditional Entropy (CE) metrics, respectively. We demonstrate that it will lead to a better latent space with high diversity and low uncertainty by controlling the distribution of posterior's  parameters across the whole data. Then, instead of introducing new loss terms or modifying training strategies,  we propose to apply Dropout~\cite{srivastava2014dropout} on the variances and Batch-Normalization on the means simultaneously to regularize their distributions implicitly. In particular, we also discuss and prove the effectiveness of two regularizations in a rigorous way. 
Furthermore, to verify the generalization of our approaches, we also demonstrate that DU-VAE can be extended empirically into VAE-IAF~\cite{kingma2016improved}, a well-known normalizing flow-based VAE.
Finally, extensive experiments have been conducted on three benchmark datasets, and the results clearly show that our approach can outperform state-of-the-art baselines on both likelihood estimation and classification tasks. 
Code and data are available at \url{https://github.com/SmilesDZgk/DU-VAE}.
\section{Background of VAE}\label{sec:pre}
Given the input space $x \in \mathcal{X}$, VAE aims to construct a smooth latent space $z \in \mathcal{Z}$ by learning a generative model $p(x,z)$. Starting from a prior distribution $p(z)$, such as standard multivariate Gaussian $\mathcal{N}(0,I)$, VAE generates data with a complex conditional distribution $p_{\theta}(x|z)$ parameterized by one neural network $f_{\theta}(\cdot)$. The goal of the model training is to maximize the marginal likelihood $E_{p_{\mathcal{D}}(x)}[\log p_{\theta}(x)]$, where the $p_{\mathcal{D}}(x)$ is the true underlying distribution.
To calculate this intractable marginal likelihood,
an amortized inference distribution $q_{\phi}(z|x)$ parameterized by one neural network $f_{\phi}(\cdot)$ has been utilized to approximate the true posterior. Then, it turns out to optimize the following lower bound:
\begin{equation}
\small
\begin{split}
\mathcal{L}= E_{p_{\mathcal{D}}(x)}[E_{q_{\phi}(z|x)}[\log p_{\theta}(x|z)]-[D_{KL}[q_{\phi}(z|x)||p(z)]],
\end{split}
\label{equ:elbo}
\end{equation}
where the first term is the reconstruction loss and the second one is the Kullback-Leibler (KL) divergence between the approximated posterior and prior.

Unfortunately, in practice, VAE may fail to capture meaningful representation.
In particular, when applying auto-regressive models as the decoder network, such as LSTMs or PixelCNN, 
it is likely to model the data marginal distribution $p_{\mathcal{D}}(x)$ very well even without latent variable $z$,
i.e., $p(x|z) = \Pi_{i}p(x_i|x_{<i})$. In this case, VAE degenerates to auto-regressive, the  latent variable $z$ tends to be independent with the data $x$.
Meanwhile, with the goal to minimize $D_{KL}[q(z|x)||p(z)]$ in ELBO objective, $q(z|x)$ vanishes to $p(z)$, 
i.e., $q(z|x_i) = q(z|x_j) =p(z)$, $\forall x_i,x_j \in \mathcal{X}$. 
To solve this problem, 
we will direct the latent space learning carefully and purposefully for high diversity and low uncertainty in the following.

\section{The Proposed Method}
Here, 
we start with theoretical analysis on the latent space of VAE from two geometric properties: diversity and uncertainty, respectively.
Then, we design Dropout on the variance parameters and Batch-Normalization on the mean parameters to encourage the latent space with high  diversity  and low uncertainty. In particular, the effectiveness of our approach will be discussed and proved rigorously.
Finally, we extend DU-VAE into VAE-IAF~\cite{kingma2016improved} empirically. 

\subsection{Geometric Properties of Latent Space}
For enabling meaningful and compact representation learning in VAE model, we have two intuitions: 1) for different data samples $x_1,x_2$, the posteriors $q(z_1|x_1)$ and $q(z_2|x_2)$ should mutually diversify  from each other, which encourages posteriors to capture the characteristic or discriminative information from data; 2) given data sample $x$, the degree of uncertainty of the latent variable $z$ should be minimized, which encourages removing  redundant information from $z$.
Guided by those intuitions, we first analyze the diversity and uncertainty of latent space under quantitative metric, respectively.

\subsubsection{Diversity of Latent Space.}  Here, we attempt to measure the divergence among the posterior distribution family.
One intuitive and reasonable metric is the expectation of the mutual divergence between a pair of posteriors. Following this idea,~\cite{ma2018mae} proposed the mutual posterior diversity (MPD) to measure the diversity of posteriors, which can be computed by:
\begin{equation}
\small
\begin{split}
MPD_{p_{\mathcal{D}}(x)}[z]& =E_{p_{\mathcal{D}(x)}}[D_{SKL}[q_{\phi}(z_1|x_1)|q_{\phi}(z_2||x_2)]],
\end{split}
\label{equ:mpd0}
\end{equation}
where $x_1,x_2\sim p_{\mathcal{D}(x)}$ are $i.i.d.$ and $D_{SKL}[q_1||q_2]$ is symmetric KL divergence defined as the mean of $D_{KL}[q_2||q_1]$ and $D_{KL}[q_2||q_1]$, which is analytical under Gaussian distributions. Specifically, we have:
\begin{equation}
\small
\begin{split}
&2MPD_{p_{\mathcal{D}}(x)}[z] =  \sum_{d=1}^n E_{p_\mathcal{D}(x)}[\frac{(\mu_{x_1,d} - \mu_{x_2,d})^2}{\delta_{x_1,d}^2}]\\
& + \sum_{d=1}^n E_{p_{\mathcal{D}(x)}}[\delta_{x,d}^2]E_{p_{\mathcal{D}(x)}}[\frac{1}{\delta_{x,d}^2}]-1.\\
\end{split}
\label{equ:mpd}
\end{equation}

Interestingly, if the value of $\delta^2_{x,d}$ is upper bounded,  like less than 1 in most practical case for  VAEs. then, we can find that  MPD has one lower and strict bound proportional to $\sum_{d=1}^n  Var_{P_{\mathcal{D}(x)}}[\mu_{x,d}]$ (see Supplementary). 

\subsubsection{Uncertainty of Latent Space.} 
Here, we aim at quantifying the uncertainty about the outcome  of latent variable $z$ given data sample $x$ and the learned  encoder distribution $q_{\phi}(z|x)$. 
In information theory, conditional entropy is utilized to measure the average level of the uncertainty inherent in the variable's possible outcomes when giving another variable. 
Due to the same goal, we follow this idea and  use the Conditional Entropy (CE) $H_{q_{\phi}}(z|x)$ for $z$ conditioned on $x$ to measure the uncertainty of latent space:
\begin{equation}
\small
\begin{split}
H_{q_{\phi}}(z|x) = E_{p_{\mathcal{D}}(x)}[H(q_{\phi}(z|x))],
\end{split}
\label{equ:ce}
\end{equation}
where  $H(q_{\phi}(z|x))$ denotes the differential entropy of posterior $q_{\phi}(z|x)$. Actually, $H(q_{\phi}(z|x))$  can be computed analytically as $\sum_{d=1}^n \frac{1}{2}\log(2 \pi e \delta^2_{x,d})$, then we have:
\begin{equation}
\small
\begin{split}
H_{q_{\phi}}(z|x) =\frac{n}{2}\log2 \pi e+  \frac{1}{2}\sum_{d=1}^nE_{p_{\mathcal{D}}(x)}[\log \delta^2_{x,d}].
\end{split}
\label{equ:entropy}
\end{equation}

Intuitively, in order to reduce the uncertainty in the latent space, we need to minimize the conditional entropy $H_{q_{\phi}}(z|x)$. However, the differential entropies $H(q_{\phi}(z|x))$ defined on continuous spaces are not bounded from below. That is, the variance $\delta^2_{x,d}$ can be scaled to be arbitrarily small achieving arbitrarily high-magnitude negative entropy. As a result, the optimization trajectories will  invariably end with garbage networks as activations approach zero or infinite.
To solve this problem, we enforce the differential entropy non-negativity by adding noise to the latent variable. For one latent variable $z$, we replace it with $z+\epsilon$, where $\epsilon \sim \mathcal{N}(0, \alpha)$ is one zero-entropy noise, where we set constant $\alpha = \frac{1}{2 \pi e }$ for convenience. Then based on properties of Gaussian distribution, we have $H(q_{\phi}(z|x)) > H(\epsilon)=0$ and $\delta^2_{x,d} > \alpha$. 

In sum, in order to encourage the diversity and decrease the uncertainty of latent space, we need to constrain both MPD in Equation~\ref{equ:mpd} and CE in Equation~\ref{equ:ce} . 
One feasible solution is to regard them as additional objectives explicitly and approximate them by using Monte Carlo in each mini-batch. However, 
the scales among different objective terms are unbalanced, which require deliberately designed normalization or careful weight parameters tuning~\cite{ma2018mae}. 

Instead, we propose control implicitly the MPD and CE without modifying the objective function.
Based on Equation~\ref{equ:mpd} and Equation~\ref{equ:ce}, we note that both MPD and CE  are only dependent on approximated  posterior' s parameters, i.e., $\mu_{x,d}$ or $\delta_{x,d}$. This inspires us to select proper regularization on the distribution of posterior' s parameters to encourage higher MPD and lower CE.
Specifically, in the following two sub-sections. 
we will introduce the application of the Dropout on variance parameters and Batch-Normalization on mean parameters respectively, and provide theoretical analysis about the effectiveness of our approach.
\subsection{Dropout on Variance Parameters}
In order to encourage high diversity and low uncertainty of latent space, we need to increase the  MPD in Equation~\ref{equ:mpd} and decrease the CE in Equation~\ref{equ:entropy}, simultaneously.
Meanwhile, we also need to avoid $E_{p_{\mathcal{D}}(x)}[\delta^2_{x,d}]$ to be too small for ensuring the smoothing of the latent space. One extreme case is that when $E_{p_{\mathcal{D}}(x)}[\delta^2_{x,d}]$ convergence to 0, i.e., $\delta^2_{x,d} \approx 0, \forall x,d$,  each data point is associated with a delta distribution in latent space and the VAEs degenerate into Autoencoders in this dimension. 
To accomplish these requirements together, we propose to apply Dropout~\cite{srivastava2014dropout} to regularize posterior's variance parameters in training as following, 
\begin{equation}
\small
\begin{split}
\hat{\delta^2}_{x,d} = g_{x,d} (\delta^2_{x,d}-\alpha) +\alpha,
\end{split}
\label{equ:dropout}
\end{equation}
where $g_{x,d}$ denotes the independent random variable generated from the normalized Bernoulli distribution $1/pB(1,p),~p \in (0,1)$, where $E_{B}[g_{x,d}]=1$. Then, we have the following proposition (see Supplementary for the proof.):
\begin{mypro}
Given the Dropout strategy defined in Equation~\ref{equ:dropout}, we have:
\begin{equation}
\small
\begin{split}
&E_{p_{\mathcal{D}}(x)\cdot B}[\hat{\delta}^2_{x,d}] = E_{p_{\mathcal{D}}(x)}[\delta^2_{x,d}], \\
&MPD_{p_{\mathcal{D}(x)}\cdot B}[z]>MPD_{p_{\mathcal{D}(x)}}[z],  \\
&H_{q_{\phi}\cdot B}(z|x)< H_{q_{\phi}}(z|x),\\
\end{split}
\label{equ:pro2}
\end{equation}
where two inequalities are both strict, the gaps between two sides are greater as $p$ decreases to 0. Then, we also have:
\begin{equation}
\small
\begin{split}
& MPD_{p_{\mathcal{D}(x)}\cdot B}[z] > \frac{1-p}{\alpha} \sum_{d=1}^n Var_{p_{\mathcal{D}}(x)}[\mu_{x,d}].\\
\end{split}
\label{equ:pro3}
\end{equation}
\label{pro:2}
\end{mypro}

Proposition~\ref{pro:2} tells us that: 1) Dropout regularization encourages the increase of $MPD_{p_\mathcal{D}(x)}[z]$ and the decrease of the conditional entropy $H_{q_{\phi}}(z|x)$ of the latent space while preserving the expectation of variance parameters, which is actually a simple but useful strategy what we need. 2) Dropout regularization also provides one lower bound of  $MPD_{p_\mathcal{D}(x)}[z]$ independent on the variance parameters, which makes it possible to ensure positive MPD with further controls on the variance $\sum_{d=1}^n Var_{p_{\mathcal{D}}(x)}[\mu_{x,d}]$. 

\subsection{Batch-Normalization on Mean Parameters} \label{sec:BN}
Inspired by Batch-Normalization (BN)~\cite{ioffe2015batch},  which is an effective approach to control the distribution of the output of neural network layer. We apply BN on the mean parameters $\mu_{x,d}$ to constrain $\sum_{d=1}^n Var_{p_{\mathcal{D}}(x)}[\mu_{x,d}]$. Mathematically, our BN is defined as:
\begin{equation}
\small
\begin{split}
\hat{\mu}_{x,d} = \gamma_{\mu_d} \frac{\mu_{x,d}-\mu_{\mathcal{B}_{d}}}{\delta_{\mathcal{B}_{d}}} + \beta_{\mu_d},
\end{split}
\end{equation}
where $\hat{\mu}_{x,d}$ represents the output of BN layer, and $\mu_{\mathcal{B}_{d}}$ and $\delta_{\mathcal{B}_{d}}$ denote the mean and standard deviation of $\mu_{x,d}$ estimated within each mini-batch.
$\gamma_{\mu_d}$ and $\beta_{\mu_d}$ are the scale and shift parameters, which lead that the distribution of $\hat{\mu}_{x,d}$ has the variance $\gamma^2_{\mu_d}$ and mean $\beta_{\mu_d}$. Therefore, we can control the $\sum_{d=1}^n Var_{p_{\mathcal{D}}(x)}[\mu_{x,d}]$ by fixing the mean $E_d[\gamma^2_{\mu_d}]=\gamma^2$ with respect to each dimension $d$. Specifically, we regard each $\gamma_d$ as learnable parameters with initialization $\gamma$. Then after each training iteration, we re-scale each parameter $\gamma_{\mu_d}$ with coefficient $\gamma / \sqrt{E_d[\gamma^2_{\mu_d}]}$.
In addition, all $\beta_{\mu_d}$ is learnable with initialization 0 and no constraint.

Overall, based on the analysis above, we propose our approach, namely DU-VAE, to encourage high diversity and low uncertainty of the latent space by applying Dropout regularizations on variance parameters and Batch-Normalization on mean parameters of approximated posteriors, simultaneously. Specifically, the training of  DU-VAE is following Algorithm~\ref{alg:1}.


\begin{algorithm}[!t]
\caption{Training Procedure of DU-VAE} \label{alg:1}
\begin{algorithmic}[1]
\STATE Initialize $\phi$, $\theta$, $\gamma_{\mu} =\gamma$, and $\beta_{\mu} = 0$
\WHILE{not convergence} 
\STATE Sample a mini-batch $x$
\STATE $\mu_{x},\delta_{x}^2 = f_{\phi}(x)$.
\STATE $\hat{\mu}_{x} = BN_{\gamma_{\mu}, \beta_{\mu}}(\mu_{x})$, $\hat{\delta}^2_{x} = Dropout_{p}(\delta_{x}^2)$.
\STATE Sample $z \sim \mathcal{N}(\hat{\mu}_{x},\hat{\delta}_{x}^2)$ and generate $x$ from $f_{\theta}(z)$.
\STATE Compute gradients $g_{\phi,\theta} \leftarrow \nabla_{\phi,\theta} \mathcal{L}_{ELBO}(x;\phi,\theta )$.
\STATE Update $\phi,\theta$, $\gamma_{\mu}$, $\beta_{\mu}$ according to $g_{\phi,\theta}$.
\STATE $\gamma_{\mu} = \frac{\gamma}{\sqrt{E[\gamma^2_{\mu}]}} \odot \gamma_{\mu}$
\ENDWHILE
\end{algorithmic}
\end{algorithm}

\noindent\textbf{Connections with BN-VAE.}
In the literature, BN-VAE~\cite{Zhu2020ABN} also applies BN on mean parameters. Zhu et al. claim that keeping one positive lower bound of the KL term, i.e., the expectation of the square of mean parameters $\sum_{d=1}^n E_{q_{\phi}}[\mu_{x,d}^2]$, is \textit{sufficient} for preventing  posterior collapse. In practice, they ensure $E_{q_{\phi}}[\mu_{x,d}^2] >0$ by fixing scale parameter $\gamma_{\mu_d}$ of BN for each dimension $d$. 
However, here, we will demonstrate that keeping one positive lower bound of MPD is one more powerful strategy for preventing collapse posterior.
As the discussion in Section~\ref{sec:pre}, when posterior collapse occurs, we have $q(z|x_i) = q(z|x_j) =p(z)$, $\forall x_i,x_j \in \mathcal{X}$. Therefore, to avoid this phenomenon, we actually need to control posterior distributions carefully so that: 
\begin{equation}
\small
\begin{split}
&q(z |x)\neq p(z),~~\exists~x \in \mathcal{X},\\
&q(z|x_i) \neq q(z|x_j),~~\exists~x_i,x_j \in \mathcal{X}.
\end{split}
\label{equ:condition}
\end{equation}
where the first term is actually  implied in the second term as the \textit{necessary} condition and $D_{KL}[q(z|x)||p(x)]>0$ is \textit{equivalent} (both \textit{sufficient} and \textit{necessary}) to the first term, we can claim that keeping one positive lower bound of the KL term is not \textit{sufficient} for the second term along with several certain abnormal cases (detailed analysis can be found in  Supplementary.). By contrast, keeping one positive MPD in the latent space is actually one \textit{equivalent} condition for the second term, which implies the first term.
Actually, from the perspective of the diversity of latent space, we can provide one more possible explanation for the effectiveness of BN-VAE. That is, the application of BN on $\mu_x$ ensures one positive value of $Var_{p_{\mathcal{D}}(x)}[\mu_x,d{}]$ for each $d$, which is also one lower bound of MPD defined in Equation~\ref{equ:mpd0} when the variance parameters has one constant upper bound, like 1 in practice.

\subsection{Extension to VAE-IAF}
Here, to further examine the generalization of DU-VAE, we aim to extend our approach for other VAE variants, such as, VAE-IAF~\cite{kingma2016improved}, one well-known normalizing flow-based VAE.
Different from classic VAEs which assume the posterior distributions are diagonal Gaussian distributions, VAE-IAF can construct more flexible posterior distributions through applying one chain of invertible transformations, named the IAF chain, on an initial random variable drown from one diagonal Gaussian distribution. 
Specifically, the initial random variable $z^0$ is sampled from the diagonal Gaussian with parameters $\mu^0$ and $\delta^0$ outputted from the encoder network. Then, $T$ invertible transformations, are applied to transform  $z^0$ into the final random variable $z^T$.
More details can be found in~\cite{kingma2016improved}.

Indeed, noting that the MPD and CE of the initial random variable $z^0$ have the same form as these for classic VAEs in Equation~\ref{equ:mpd0} and Equation~\ref{equ:ce}, one intuitive idea is to apply Dropout on $\delta^0$ and  Batch Normalization on $\mu^0$ with the guidance in Algorithm~\ref{alg:1} to control the MPD and CE of $z^0$. It is surprising to find that this simple extension of DU-VAE, called DU-IAF, demonstrated comparative performance in our experiments. This may be attributed to the close connection between $z^0$ and $z^T$.
In particular, we find that the CE of $z^0$ is the upper bound of CE of $z^T$. Meanwhile, $MPD_{p_{\mathcal{D}}(x)}[z^0]$ is closely related with $MPD_{p_{\mathcal{D}}(x)}[z^T]$, even they are equal to each other when each invertible transformation in IAF chain is independent on the input data. Further discussion and proof can be found in Supplementary.

\begin{table*}[!t]
\center
\footnotesize
\setlength{\tabcolsep}{2.6mm}{
\begin{tabular}{ccccccccccccc}\hline \hline
                 & \multicolumn{4}{c}{Yahoo}       & \multicolumn{4}{c}{Yelp}       & \multicolumn{4}{c}{OMNIGLOT}        \\ \hline
Model            & NLL & KL  & MI   &AU  & NLL  & KL  & MI & AU & NLL  & KL  & MI & AU     \\ \hline
VAE             &328.5&0.0 &0.0 &5.0 &357.5&0.0 &0.0 &0.0 &89.21&2.20 &2.16 &5.0  \\
$\beta$-VAE*{\scriptsize(0.4/~0.4/~0.8)} &328.7&6.4 &6.0 &13.0 &357.4&5.8 &5.6 &4.0 &89.15&9.98 &3.84 &13.0  \\
SA-VAE*           &327.2&5.2 &2.7 &8.6 &355.9&2.8 &1.7 &8.4 &89.07&3.32 &2.63 &8.6  \\
Agg-VAE          & \textbf{326.7} &5.7 &2.9 &6.0 &355.9&3.8 &2.4 &11.3 &89.04&2.48 &2.50 &6.0  \\
FB {\scriptsize (0.1)}           &328.1&3.4 &2.5 &32.0 &357.1&4.8 &2.5 &32.0 &89.17&7.98 &6.87 &32.0  \\
$\delta$-VAE {\scriptsize(0.1) }   &329.0&3.2 &0.0 &2.0 &357.6&3.2 &0.0 &0.0 &89.62&3.20 &2.36 &2.0  \\
BN-VAE {\scriptsize(0.6/~0.6/~0.5) }     &326.9&8.3 &7.0 &32.0 &355.7&6.0 &5.2 &32.0 &89.26&4.34 &4.03 &32.0  \\
MAE {\scriptsize(1/~2/~0.5, 0.2/~0.2/~0.2)}        &332.1&5.8 &3.5 &28.0 &362.8&8.0 &4.6 &32.0 &89.62&15.61 &8.90 &32.0  \\ \hline
DU-VAE {\scriptsize(0.5, 0.9)}         &327.0&5.2 &4.3 &18.0 &355.6&5.3 &4.9 &18.0 &  \textbf{89.00} &6.63 &5.97 &19.0  \\
DU-VAE {\scriptsize(0.5, 0.8)}         &327.0&6.7 &6.0 &19.0 &  \textbf{355.5 } &6.8 &5.9 &18.0 &89.04&7.46 &6.31 &32.0  \\
DU-VAE {\scriptsize (0.6, 0.8)}           &  \textbf{326.7} &8.7 &7.2 &28.0 &355.8&9.6 &7.7 &23.0 &89.18&10.99 &8.22 &32.0  \\ \hline\hline
IAF+FB {\scriptsize(0.15/0.25/0.15)}        & 328.4    & 5.2    & -   & -  & 357.1 &7.7    &-   & -    &88.98&6.77 &- &- \\
IAF+BN {\scriptsize(0.6/0.7/0.5)}        & 328.1    & 0.2   & - & -   & 356.6    & 0.6    & -   & -    &89.32&1.30 &- &-  \\\hline
DU-IAF {\scriptsize(0.7/0.6/0.5, 0.70/0.70/0.85)}         & \textbf{327.4}    & 5.4    & -    & -    & \textbf{356.1}    & 5.1    & -    & -   &\textbf{88.97} &6.77 &- &- \\
\hline\hline
\end{tabular}}
\caption{The performance on likelihood estimation. Due to the intractability of MI and AU metrics for IAF-based models, we just report NLL and KL same as \protect \cite{kingma2016improved}. * indictes the results are referred from \protect \cite{he2018lagging}. Hyper-parameters are reported in brackets and split by slashes  if different on different datasets.} 
\label{tab:nll1}
\end{table*}

\section{Experiments}
In this section, 
our method would be evaluated on three benchmark datasets in terms of various metrics and tasks. 
Complete experimental setup can be found in Supplementary.

\subsection{Experimental Setup}
\noindent\textbf{Setting.} Following the same configuration  as~\cite{he2018lagging}, we evaluated our method on two text benchmark datasets, i.e., Yahoo and Yelp corpora~\cite{yang2017improved} and one image benchmark dataset, i.e., OMNIGLOT~\cite{lake2015human}. 
For text datasets, we utilized a single layer LSTM as both encoder and decoder networks, where the initial state of the decoder is projected by the latent variable $z$.
For images, a 3-layer ResNet~\cite{he2016deep} encoder and a 13-layer Gated PixelCNN~\cite{van2016conditional} decoder are applied.
We set the dimension of $z$ as 32.
and utilized SGD to optimize the ELBO objective for text and Adam~\cite{kingma2014adam} for images. 
Following~\cite{burda2016importance}, we utilized dynamically binarized images for training and  the fixed binarization as test data.
Meanwhile, following~\cite{bowman2015large}, we applied a linear annealing strategy to increasing the KL weight from 0 to 1 in the first 10 epochs if possible.

\noindent\textbf{Evaluation Metrics.}
Following~\cite{burda2016importance}, we computed the approximate negative log-likelihood (NLL) by estimating 500 importance weighted samples.
In addition, 
we also considered the value of  KL term, mutual information (MI) $I(x,z)$~\cite{alemi2016deep} under the joint distribution $q(x,z)$ and the number of activate units (AU)~\cite{he2018lagging}  as additional metrics.
In particular, 
the activity of each dimension $z_d$ is measured as $A_{z,d} = Cov(E_{z_d \sim q(z_d|x)}[z_d])$. One dimension is regarded to be active when $A_{z,d}>0.01$.

\noindent\textbf{Baselines.} 
We compare our method with various VAE-based models, which can be grouped into two categories: 
1) Classic VAEs:
\textbf{VAE} with annealing~\cite{bowman2015large};
 Semi-Amortized VAE (\textbf{SA-VAE})~\cite{kim2018semi};
\textbf{Agg-VAE}~\cite{he2018lagging};
\textbf{$\beta$-VAE}~\cite{Higgins2017betaVAELB} with parameter $\beta$ re-weighting the KL term;
\textbf{FB}~\cite{kingma2016improved} with parameter $\lambda$ constraining the minimum of KL~term in each dimension;
 \textbf{$\delta$-VAE}~\cite{razavi2018preventing} with parameter $\delta$ constraining the range of KL term;
 \textbf{BN-VAE} ~\cite{Zhu2020ABN} with parameter $\gamma$ keeping one positive KL term;
 \textbf{MAE}~\cite{ma2018mae} with parameters $\gamma$ and $\eta$ controlling the diversity and smoothness of the latent space. Note that we implemented MAE with the standard Gaussian prior, instead of the AF prior in~\cite{ma2018mae} for one fair comparison.
2) IAF-based models: 
 \textbf{IAF+FB}~\cite{kingma2016improved}, which utilized the FB strategy with the parameter $\lambda$ to avoid the posterior collapse in VAE-IAF;
 \textbf{IAF+BN}, where we applied BN regularization on the mean parameters of the distributions of $z^0$ with the fixed scale parameters $\gamma$ in each dimension.

\subsection{Overall Performance}
\noindent \textbf{Log-Likelihood Estimation.}
Table~\ref{tab:nll1}  shows the results in terms of log-likelihood estimation. We can note that DU-VAE and DU-IAF achieve the best NLL among classic VAEs and IAF-based VAEs in all datasets, respectively.
Besides, we also have some interesting findings.
First, MAE does not perform well in all datasets, which may be caused by the difficulty to balance the additional training objective terms and ELBO.
Second, although,  Agg-VAE and SA-VAE also  reach the great NLL in both  datasets, they require the additional training procedure on the inference network, leading to the high training time cost~\cite{Zhu2020ABN}.
Third, BN-VAE also achieves completive performance on text datasets. However, for images, where the posterior collapse may be less of an issue~\cite{kim2018semi}, BN-VAE fails to catch up with other models, even worse than basic VAE on NLL. Fourth, DU-VAE prefers to capture higher KL and MI compared with BN-VAE with the same scale parameter $\gamma$. In other words,  DU-VAE can convert more information from the observed data into the latent variable. 
Fifth, based on the results of IAF+BN, we can find that the BN strategy used in BN-VAE can not prevent the collapse posterior in VAE-IAF with small KL. By contrast, our approach can be easily extended for VAE-IAF with the best performance.
Finally, we also note that IAF based models may be more suitable for image dataset without sound performance on text, while DU-IAF nevertheless achieves competitive performance.

\begin{table}[!t]
\center
\footnotesize
\begin{tabular}{cccccc}\hline\hline 
$\#$label       & 100  & 500  & 1k & 2k & 10k \\\hline
AE           & 84.05 & 86.82 & 87.93 & 88.19 & 88.75  \\
VAE          & 71.10 & 71.43 & 71.58 & 72.96 & 77.11\\
$\delta$-VAE {\scriptsize(0.1) }& 60.11& 60.52 &61.46& 63.79 & 64.38  \\
Agg-VAE       & 75.05 &77.16 &  78.50  &  79.29  &  80.07 \\
FB {\scriptsize(0.1) }       &75.19 & 80.78 & 81.63 & 82.28 & 82.39 \\
BN-VAE{\scriptsize (0.6) }     & 84.53   &88.22  &89.45 &89.63 & 89.72 \\
MAE {\scriptsize(2, 0.2)}        & 61.50 & 61.70 & 62.42&63.58 &63.68  \\
 \hline
DU-VAE {\scriptsize (0.5, 0.8)}      &  \textbf{88.91}&\textbf{89.63}&\textbf{90.36}&\textbf{90.51}&\textbf{90.77} \\ 
\hline\hline
IAF+FB{\scriptsize(0.25)}&89.73&90.60&90.94&90.91 &91.01\\
IAF+BN{\scriptsize(0.7)} &87.98&89.03& 89.18&89.35 &90.29\\ \hline
DU-IAF {\scriptsize(0.6, 0.7)}&\textbf{91.25}&\textbf{91.10}&\textbf{91.52}&\textbf{91.97}&\textbf{92.31}\\
 \hline\hline  
\end{tabular}
\caption{The accuracy of the classification on Yelp.}
\label{tab:cla1}
\vspace{-2mm}
\end{table}

\begin{table}[!t]
\center
\footnotesize
\setlength{\tabcolsep}{3.7mm}{
\begin{tabular}{cccc}\hline\hline 
$\#$label for each character      & 5  & 10  & 15 \\\hline
AE&37.28&43.38&46.94\\
VAE&29.48&37.79&42.24\\
$\delta$-VAE {\scriptsize(0.1)}&37.28&43.38&46.94\\
Agg-VAE&33.72&41.31&46.27\\
FB {\scriptsize(0.1)} &33.93&41.05&45.21\\
BN-VAE {\scriptsize(0.5)} &31.17&39.15&43.24\\
MAE {\scriptsize(05, 0.2)} &35.05&41.72&44.95\\ \hline
DU-VAE {\scriptsize(0.5, 0.1)}&\textbf{40.54}&\textbf{48.09}&\textbf{52.47}\\ \hline\hline
IAF+FB{\scriptsize(0.15)}&38.33&45.85&49.90\\
IAF+BN{\scriptsize(0.5)} &16.58&19.49&21.11\\ \hline
DU-IAF {\scriptsize(0.5, 0.15)}&\textbf{41.84}&\textbf{49.86}&\textbf{52.97}\\
 \hline\hline  
\end{tabular}}
\caption{The average accuracy of classifications on  OMNIGLOT.}
\label{tab:cla2}
\vspace{-2mm}
\end{table}


\begin{figure*}[!t]
\centering
\subfigure[{\scriptsize True Latent Space}]{ 
\includegraphics[width=0.27\columnwidth]{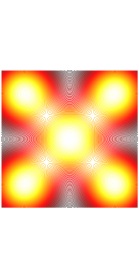}} \hspace{0.03\columnwidth}
\subfigure[{\scriptsize VAE}]{ 
\includegraphics[width=0.27\columnwidth]{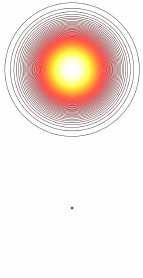}} \hspace{0.03\columnwidth}
\subfigure[{\scriptsize Agg-VAE}]{ 
\includegraphics[width=0.27\columnwidth]{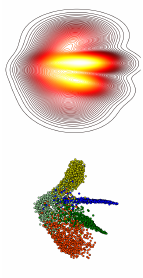}} \hspace{0.03\columnwidth}
\subfigure[{\scriptsize BN-VAE} {\scriptsize (1.0)}]{ 
\includegraphics[width=0.27\columnwidth]{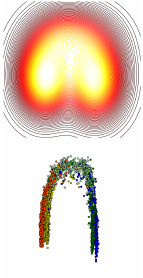}} \hspace{0.03\columnwidth}
\subfigure[{\scriptsize DU-VAE} {\scriptsize (1.0, 0.5)}]{ 
\includegraphics[width=0.27\columnwidth]{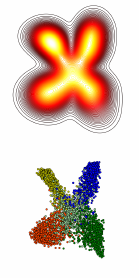}}
\caption{The visualization of the latent space learned by DU-VAE and other baselines.
Figure (a) is the counter plot of the true latent space for generating the synthetic dataset.
In the rest,
the first line shows the counter plot of the aggregated posterior $q_{\phi}(z)$. The brighter the color, the higher the probability.
Meanwhile, the location of mean parameters  are displayed in the second line with colors to distinguish different categories generated from different Gaussian components, where the blue ones correspond to the component in the center in Figure (a), and the others denote the other four components.
All figures are located in the same region, i.e., $z \in [-3,3]\times [-3,3]$, with the same scale.
}
\label{fig:2}
\vspace{-3mm}
\end{figure*}

\noindent\textbf{Classification.}
To evaluate the quality of learned representation, we train a one-layer linear with the output from the trained model as the input for classification tasks on both text and image datasets. 
For classic VAEs, the mean parameter $\mu$ of each latent variable has been used as the representation vector. For IAF-based models, we first selected the initial sample $z^0$ in latent space as its mean parameter $\mu^0$. Then, the combination of $z^0$ and $z^T$ is used as the representation vector.

Specifically, for text datasets, following~\cite{shen2017style}, we work with one downsampled version of Yelp sentiment dataset for binary classification.
Table~\ref{tab:cla1}  shows the performance under varying  number of labeled data. 
For the image dataset, noting that OMNIGLOT contains 1623 different handwritten characters from 50 different alphabets, where each character has 15 images in our training data and 5 images in our testing data, we conducted classifications on each alphabet with varying number of training samples for each character.
Table~\ref{tab:cla2} reports the average accuracy. 

We can find that DU-VAE and DU-IAF achieve the best accuracy under all settings for classic VAEs and IAF-based models respectively. Interestingly, we also find that most baselines show inconsistent results on text and image  classification. For example, Agg-VAE and BN-VAE may be better at text classification without sound accuracy in Table~\ref{tab:cla2}. On the contrary, $\delta$-VAE and MAE adapt to image classification better with uncompetitive performance in Table~\ref{tab:cla1}.
Meanwhile, we note IAF chain trends to improve the classification accuracy for FB and our approach to both text and image datasets. However, IAF+BN fails to achieve competitive performance on image classification, which indicates that the applications of BN in BN-VAE may not be suitable for image again.

\begin{figure}[!t]
\centering
\subfigure[{\scriptsize Yahoo}]{ 
\includegraphics[width=0.30\columnwidth]{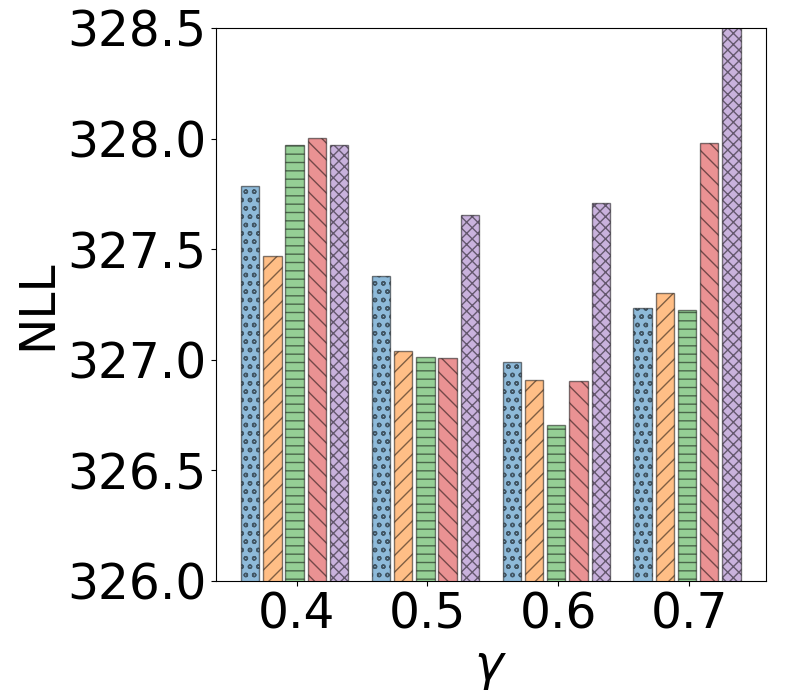} }
\subfigure[{\scriptsize Yelp}]{ 
\includegraphics[width=0.30\columnwidth]{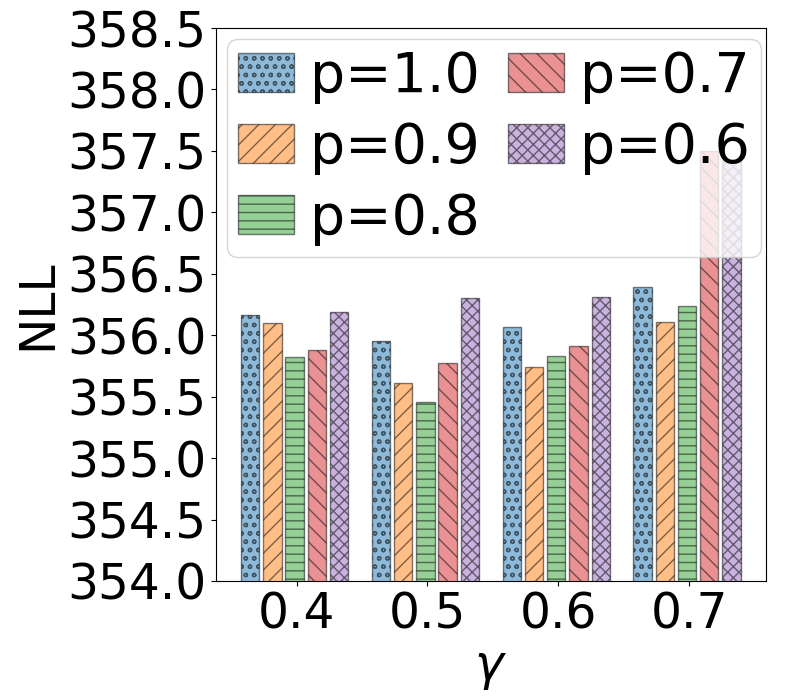} }
\subfigure[{\scriptsize OMNIGLOT}]{ 
\includegraphics[width=0.30\columnwidth]{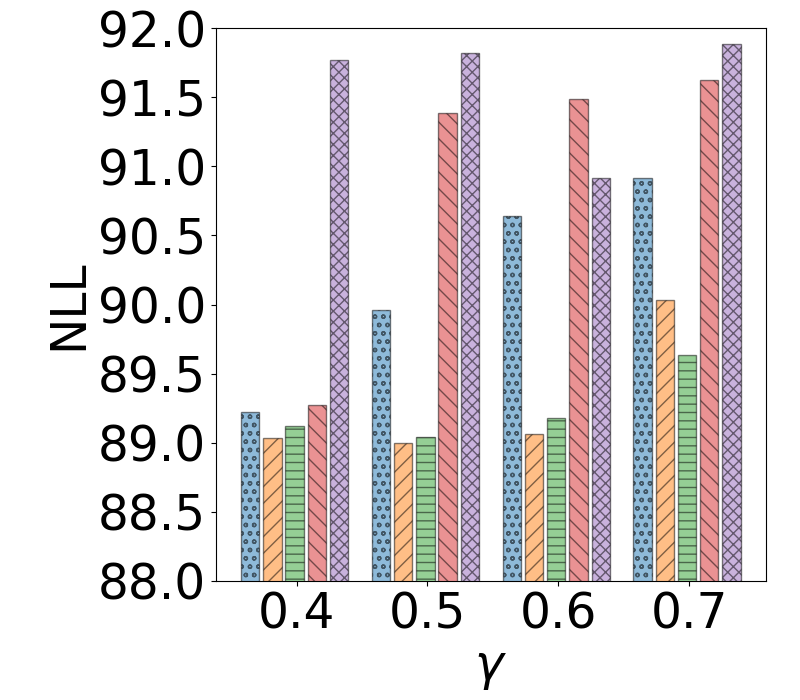} }
\caption{Parameter Analysis.}
\label{fig:1}
\vspace{-3mm}
\end{figure}

\noindent \textbf{Parameter Analysis.} Here, we train DU-VAE by varying $\gamma$ from 0.4 to 0.7 and $p$ from 1 to 0.6. 
As the Figure~\ref{fig:1} shows,
we find that, DU-VAE would achieve the best NLL  with parameters ($\gamma$, $p$) as when ($0.6$, $p=0.8$)  for Yahoo,  ($0.5$, $p=0.8$) for Yelp,  and  ($0.5$, $p=0.9$) for OMNIGLOT, respectively.

\vspace{-1mm}
\subsection{Case Study--Latent Space Visualization}
Here, we aim to provide one intuitive comparison of latent spaces learned by different models based on one simple synthetic dataset. Specifically, following~\cite{kim2018semi}, we first sample 2-dimensional latent variable $z$ from one mixture of Gaussian distributions that have 5 mixture components. Then one text-like dataset can be generated from one LSTM layer conditioned on those latent variables.
Based on this synthetic dataset, we trained different VAEs  with 2-dimensional standard Gaussian prior and diagonal Gaussian posterior. Then, we visualize the learned latent spaces by displaying the counter plot of  the aggregated approximated posteriors $q(z)=E_{p_{\mathcal{D}}(x)}[q_{\phi}(z|x)]$  and  the location of approximated posterior's mean parameters for different samples $x$.  

According to the results in Figure~\ref{fig:2}, we have some interesting observations.
First, due to the posterior collapse, VAE learns an almost meaningless latent space where  the posterior $q(z|x)$ for all data are squeezed  in the center.  
Actually, it is not surprising that the aggregated posterior  matches  the prior excessively in this case, because we almost have  $q_{\phi}(z|x)=p(z)$, $\forall x$.
Second,  Agg-VAE, BN-VAE, and DU-VAE all tend to diverse samples in different categories, but in different manners and degrees. 
Intuitively, all three models force to embedding the blue category in the center around by the other four categories. However, only the average posterior learned by DU-VAE have five centers same as the true latent space. 
Meanwhile, 
DU-VAE with Dropout strategy encourages the aggregated  posteriors to be more compact, while that of BN-VAE is more broad, compared with the prior. 
Those observations demonstrate that  DU-VAE  tends to guide the latent space to be more diverse and less uncertain.

\vspace{-2mm}
\section{Conclusion}
In this paper, we developed a novel generative model, DU-VAE, for learning a more diverse and less uncertain latent space. The goal of DU-VAE is to ensure that more  meaningful and compact representations can be learned. Specifically, we first demonstrated theoretically that it led to better latent space with high diversity and low uncertainty awareness by controlling the distribution of posterior's  parameters across the whole dataset respectively. Then, instead of introducing new loss terms or modifying training strategies, we proposed to apply Dropout on the variances and Batch-Normalization on the means simultaneously to regularize their distributions implicitly. 
Furthermore,
we extended DU-VAE into VAE-IAF empirically.
The experimental results on three benchmark datasets clearly showed that DU-VAE  outperformed state-of-the-art baselines on both likelihood estimation and underlying classification tasks. 

\vspace{-2mm}
\section*{Acknowledgements}
This work was partially supported by grants from the National Natural Science Foundation of China (Grant No. 91746301, 61836013).


\bibliographystyle{named}
\bibliography{mybib}
\appendix
\section*{Supplementary}
\makeatletter
\setcounter{table}{0}  
\setcounter{figure}{0}
\setcounter{algorithm}{0}
\renewcommand{\thefigure}{S\@arabic\c@figure}
\renewcommand{\thetable}{S\@arabic\c@table}
\renewcommand{\theequation}{S\@arabic\c@equation}
\renewcommand{\thealgorithm}{S\@arabic\c@algorithm}
\makeatother

\section{Related Work}
Representation learning as one important direction of machine learning has attracted enormous attention and severed for various applications~\cite{blei2003latent,huai2014topic,zhu2014tracking,lin2017collaborative,xu2018measuring,shen2018joint,radford2015unsupervised}, Among them, Variational Autoencoder (VAE)~\cite{kingma2013auto} have achieve the great success in recent years.
The literature for enhancing the representation learning of VAE can be roughly divided into two categories based on different motivations: solving optimization challenge and directing latent space.

\noindent\textbf{Solving Optimization Challenge.} \cite{bowman2015large} first detailed that VAE tends to ignore the latent variable when employing one LSTM as the decoder network.
They interpreted this problem as optimization challenges of VAE. Following this idea, various strategies have been proposed to spur the training trajectory  jump out from the local optima, namely, \textit{posterior collapse}.

One direction is to adjust the training strategy.
For example, \cite{bowman2015large} proposed KL annealing to address this problem, by slightly increasing the weight of KL term during the first few epochs of training. 
\cite{kim2018semi} designed Semi-Amortized VAE (SA-VAE) to compose the inference network with additional updates.
More recently, ~\cite{he2018lagging} proposed to aggressively optimize inference network multiple times before a single decoder update. 
However, those approaches often suffer from the additional training procedure.
Meanwhile, weakening the expressive capacity of decoder networks is another option.
~\cite{semeniuta2017hybrid} and  ~\cite{yang2017improved} implemented the decoder by CNN without autogressive modeling.
~\cite{Chen2017VariationalLA} applied a lossy representation as the input of autogressive decoder to force the latent representation to discard irrelevant information.
However, those approaches are often  problem-specific and require manual designs of the decoder network.

Some other works constraint the minimum of the KL terms to prevent the KL from vanishing to 0. 
For instance,  $\beta$-VAE~\cite{Higgins2017betaVAELB} introduces  a hyperparameter to weight the KL term that constrains its minimum.
  Free bits (FB) \cite{kingma2016improved} replaces the KL terms with a hinge loss term that maximizes the original KL with a constant.  
$\delta$-VAE~\cite{razavi2018preventing}  constrains the variational family carefully such that the posterior can never exactly recover the prior.
However, those approaches often suffer from performance degradation in likelihood estimation or no-smooth objective~\cite{Chen2017VariationalLA}.
Recently, BN-VAE~\cite{Zhu2020ABN} directly utilizes the Batch-Normalization~\cite{ioffe2015batch} on mean parameters to keep one positive lower bound of the KL term, which has witnessed a surprising effectiveness on preventing posterior collapse.
But the theoretical basis behind the effect of BN on latent space  is not yet understood. In our paper, we will provide one more possible explanation based on the geometry analysis of latent space.

\noindent\textbf{Directing Latent Space Learning.} Some other studies attempt to enhance the representation learning by directing the latent space learning.
One common idea is to use the additional mutual information-based objective terms to enforce the relationship between the latent variable and observed data. However, since the MI term can not be computed analytically, auxiliary networks~\cite{fang2019implicit,rezaabad2020learning},  Monte-Carlo~\cite{hoffman2016elbo}, or Maximum-Mean Discrepancy method~\cite{dziugaite2015training,Zhao2019InfoVAEBL} are required for computing the approximation, which introduces additional training effort or expensive computation.
Recently, MAE ~\cite{ma2018mae} further proposed Mutual Posterior-Divergence (MPD) regularization to control the geometry diversity of the latent space, which has one similar goal with MI to measure the average of divergence between each conditional  distribution and  marginal distribution. However, the scales of MPD and the original objective of VAE are unbalanced, which requires additional deliberated normalization.
In our paper, we also aim at directing the latent space learning and encouraging diversity. Different from MAE, we control MPD by network regularizations, i.e., Dropout and Batch-Normalization, without adding explicit training objectives.

\section{Proof of Equation 3}
\begin{proof}
First, we can derive the formula of $D_{SKL}[q_{\phi}(z_1|x_1)||q_{\phi}(z_2|x_2)]$ under Gaussian distributions as following:
\begin{equation}
\small
\begin{split}
&4D_{SKL}[q_{\phi}(z_1|x_1)||q_{\phi}(z_2|x_2)] = \\
&=(\mu_{x_1,d}-\mu_{x_2,d})^2(\frac{1}{\delta_{x_1,d}^2} +\frac{1}{\delta_{x_2,d}^2}) + \frac{\delta^2_{x_1,d}}{\delta^2_{x_2,d}}+\frac{\delta^2_{x_2,d}}{\delta^2_{x_1,d}} -2. \\
\end{split}
\end{equation}
 
Then, we compute the expectation of both sides of the equality with respect to $i.i.d.~x_1,x_2 \sim P_{\mathcal{D}}(x)$, we have the Equation 3.

In addition, when the value of $\delta_{x,d}^2$ is upper bound with constant $C$, we further have:
\begin{equation}
\small
\begin{split}
&4D_{SKL}[q_{\phi}(z_1|x_1)||q_{\phi}(z_2|x_2)] = \\
&\geq (\mu_{x_1,d}-\mu_{x_2,d})^2(\frac{1}{\delta_{x_1,d}^2} +\frac{1}{\delta_{x_2,d}^2})\\
&\geq (\mu_{x_1,d}-\mu_{x_2,d})^2\frac{2}{C}
\end{split}
\end{equation}
where the equality in the second line holds when $\delta_{x_1,d}^2 = \delta_{x_2,d}^2$.
Then, we compute the expectation of both sides of the inequality, we have:
\begin{equation}
\small
\begin{split}
&MPD_{p_{\mathcal{D}}(x)}[z]\geq \frac{1}{C} \sum_{d=1}^n Var_{p_{\mathcal{D}(x)}}[\mu_{x, d}],~~\text{if}~~\forall~\delta^2_{x,d}\leq C.
\end{split}
\end{equation}
\end{proof}

\section{Proof of Proposition 1}
\begin{proof}
Given the Dropout regularization defined in our paper, it is not hard  to derive the following equations:
\begin{equation}
\small
\begin{split}
&E_{B}[\hat{\delta}^2_{x,d}] = \delta^2_{x,d}, \\
&E_{B}[\frac{1}{\hat{\delta}^2_{x,d}}] = \frac{p^2}{\delta_{x,d}^2  + (p-1)\alpha} + \frac{1-p}{\alpha},\\
&E_{B}[\log \hat{\delta}^2_{x,d}]= p\log(\frac{\delta_{x,d}^2 + (p-1)\alpha}{p\alpha})  +\log \alpha.\\
\end{split}
\label{equ:pro2_1}
\end{equation}

Interestingly, we note that the second and third equations are strictly increasing and decreasing , respectively, as  $p$ decreases from 1 to 0, which can be verified by analyzing their derivatives. Then, we have:
\vspace{-1mm}
\begin{equation}
\small
\begin{split}
&\frac{1}{\alpha} > E_{B}[\frac{1}{\hat{\delta}^2_{x,d}}] > \frac{1}{\delta_{x,d}^2},\\
&\log \alpha < E_{B}[\log \hat{\delta}^2_{x,d}] < \log \delta_{x,d}^2.\\
\end{split}
\label{equ:pro2_2}
\vspace{-1mm}
\end{equation}

Meanwhile, by considering $MPD_{p_{\mathcal{D}}(x) \cdot B}[z]$ and $H_{q_{\phi}\cdot B}(z|x)$ as the function of $E_{B}[\frac{1}{\hat{\delta}^2_{x,d}}] $ and $E_{B}[\log \hat{\delta}^2_{x,d}] $ respectively, we can found they are also increasing and decreasing monotonously as $p$ decreases. Then, we turn to prove the three inequalities in Equation  7 and 8.

First, by computing the expectation of the second inequality with respect to $x \sim p_{\mathcal{D}}(x)$, we can derive:
\begin{equation}
\small
\begin{split}
&E_{p_{\mathcal{D}}(x) \cdot B}[\log \hat{\delta}^2_{x,d}]<E_{p_{\mathcal{D}}(x)}[\log \delta^2_{x,d}].
\end{split}
\label{equ:pro2_3}
\end{equation}

Meanwhile, based on Equation 5, we have:
\begin{equation}
\small
\begin{split}
&H_{q_{\phi}\cdot B}(z|x) < H_{q_{\phi}}(z|x).
\end{split}
\label{equ:pro2_4}
\end{equation}

In addition, due to the general observation:
\begin{equation}
\small
\begin{split}
&E_{p_{\mathcal{D}}(x)\cdot B}[\frac{(\mu_{x_1,d}-\mu_{x_2,d})^2}{\hat{\delta}^2_{x_1,d}} ]\\
&=E_{p_{\mathcal{D}}(x)}[(\mu_{x_1,d}-\mu_{x_2,d})^2E_{B}[\frac{1}{\hat{\delta}^2_{x_1,d}}]],
\end{split}
\end{equation}
we have the following derivation:
\begin{equation}
\small
\begin{split}
&E_{p_{\mathcal{D}}(x)\cdot B}[\frac{(\mu_{x_1,d}-\mu_{x_2,d})^2}{\hat{\delta}^2_{x_1,d}} ] < \frac{2}{\alpha} Var_{p_{\mathcal{D}}(x)}[\mu_{x,d}],\\
&E_{p_{\mathcal{D}}(x)\cdot B}[\frac{(\mu_{x_1,d}-\mu_{x_2,d})^2}{\hat{\delta}^2_{x_1,d}} ]> E_{p_{\mathcal{D}}(x)}[\frac{(\mu_{x_1,d} - \mu_{x_2,d})^2}{\delta^2_{x_1,d}}].
\end{split}
\vspace{-1mm}
\end{equation}
Then, combining Equation 3, S4, S5 and S9, we have:
\begin{equation}
\small
\begin{split}
&2MPD_{p_{\mathcal{D}}(x)\cdot B}[z] =  \sum_{d=1}^n E_{p_{\mathcal{D}}(x)\cdot B}[\frac{(\mu_{x_1,d} - \mu_{x_2,d})^2}{\hat{\delta}_{x_1,d}^2}]\\
& + \sum_{d=1}^n E_{p_{\mathcal{D}}(x) \cdot B}[\hat{\delta}_{x,d}^2]E_{p_{\mathcal{D}}(x)\cdot B}[\frac{1}{\hat{\delta}_{x,d}^2}]-1\\
&> \sum_{d=1}^n E_{p_{\mathcal{D}}(x)}[\frac{(\mu_{x_1,d} - \mu_{x_2,d})^2}{\delta_{x_1,d}^2}]\\
& + \sum_{d=1}^n E_{p_{\mathcal{D}(x)}}[\delta_{x,d}^2]E_{p_{\mathcal{D}}(x)}[\frac{1}{\delta_{x,d}^2}]-1 =2MPD_{p_{\mathcal{D}}(x)}[z].
\end{split}
\vspace{-1mm}
\end{equation}

Meanwhile, we note that  $E_{B}[\frac{1}{\hat{\delta}^2_{x,d}}] > \frac{1-p}{\alpha}$  based on Equation~\ref{equ:pro2_1}, so we also have:
\begin{equation}
\small
\begin{split}
&MPD_{p_{\mathcal{D}}(x)\cdot B}[z] > \\
&\sum_{d=1}^n E_{p_{\mathcal{D}}(x)\cdot B}[\frac{(\mu_{x_1,d}-\mu_{x_2,d})^2}{2\hat{\delta}^2_{x_1,d}} ]> \frac{1-p}{\alpha} \sum_{d=1}^n Var_{p_{\mathcal{D}}(x)}[\mu_{x,d}].\\
\end{split}
\label{equ:pro3_1}
\end{equation}
\end{proof}

\section{Connection between $z^0$ and $z^T$ in VAE-IAF}
Here, we first introduce the background knowledge for VAE-IAF.  Then we will discuss the strong relation between $z^0$ and $z^T$ in VAE-IAF with respect to CE and MPD, repspectively, which provide one explanation for the effectiveness of DU-IAF.
In addition, Algorithm~\ref{alg:2} shows the detailed training procedure of DU-IAF.

\noindent\textbf{background knowledge of VAE-IAF.} 
VAE-IAF aims to construct more flexible and expressive posterior distributions with the help of normalizing flows.
Here, we briefly introduce the main steps on constructing posterior distribution in VAE-IAF.
In practice, first, the encoder network outputs $\mu^0$ and  $\delta^0$,  in addition to an extra embedding of the input data $h$.
Then, the initial random sample can be drown from the diagonal Gaussian $q_{\phi}(z^0|x) = \mathcal{N}(\mu^0, (\delta^0)^2)$.
Second,  one chain of nonlinear invertible transformations has been defined with $T$ of IAF blocks:
\begin{equation}
\small
\begin{split}
&(m^t,s^t) = \text{AutoregressiveNeuralNet}_t(z^{t-1}, h; \psi_t),\\
&\delta^t = \text{sigmod}(s^t),
z^t = \delta^t \odot z^{t-1} + (1-\delta^t) \odot m^t,
\end{split}
\label{equ:IAF}
\end{equation}
where each IAF block is one different autoregressive neural network, which is structured to be autoregressive w.r.t. $z^{t-1}$. Finally, the final iterate $z^T$ is considered as the posterior approximate and fed into the decoder network. 
 As a result, $\frac{dz^{t}}{dz^{t-1}}$ is triangular with $\delta^t$ on the diagonal. 
In other words, all transformations are invertible with positive determinant $\prod_{d=1}^D \delta^t_d$ everywhere.


\noindent\textbf{$H_{q_{\phi}}(z^0|x)$ VS $H_{q_{\phi}}(z^T|x)$.} Here, we will prove that the CE of $z^0$ is one upper bound of that of $z^T$, i.e., $H_{q_{\phi}}(z^0|x) >H_{q_{\phi}}(z^T|x)$.
\begin{proof}
Specifically, we first need to compute the differential entropy $H(q(z^T|x))$ of posterior $q(z^T|x)$ given $x$:
\vspace{-1mm}
\begin{equation}
\small
\begin{split}
&H(q(z^T|x)) = -\int q(z^T|x) \log q(z^T|x) dz^T\\
&=- \int q(z^0|x) \det\left|\frac{dz^0}{dz^T} \right| \log q(z^0|x) \det\left|\frac{dz^0}{dz^T} \right| dz^T\\
&=-\int q(z^0|x)\log q(z^0|x) dz^0 - \int q(z^0|x)\log \det\left|\frac{dz^0}{dz^T} \right| dz^0\\
&= H(q(z^0|x)) + E_{q(z^0|x)}[\sum_{d=1}^D \sum_{t=1}^T \log \delta^t_{x,d}],
\end{split}
\vspace{-1mm}
\end{equation}
where the quality arises because each step in  IAF chain defined in Equation~\ref{equ:IAF} is invertible and differentiable, and the Jacobian matrix $\frac{dz^T}{dz^0}$ is inverse with the following determinant everywhere:
\vspace{-1mm}
\begin{equation}
\small
\begin{split}
\det |\frac{dz^T}{dz^0}| = \prod_{t=1}^T \det\left| \frac{dz^t_d}{dz^{t-1}_d}\right| = \prod_{t=1}^T\prod_{d=1}^D \delta^t_d .
\end{split}
\vspace{-1mm}
\end{equation}

Meanwhile, noting that each $\delta^t_d<1$, we have:
\vspace{-1mm}
\begin{equation}
\small
\begin{split}
H(q(z^0|x))>H(q(z^T|x)).
\end{split}
\vspace{-1mm}
\end{equation}

Then, by computing the expectation of both with respect to $x \sim p_{\mathcal{D}}(x)$, we can drive that $H_{q_{\phi}}(z^0|x) >H_{q_{\phi}}(z^T|x)$.
\end{proof}

\begin{algorithm}[!t]
\caption{Training Procedure of DU-IAF} \label{alg:2}
\begin{algorithmic}[1]
\STATE Initialize $\phi$, $\theta$, $\gamma_{\mu} =\gamma$, and $\beta_{\mu} = 0$
\WHILE{not convergence} 
\STATE Sample a mini-batch $x$
\STATE $\mu^0_{x},(\delta_{x}^0)2, h = f_{\phi}(x)$.
\STATE $\hat{\mu}^0_{x} = BN_{\gamma_{\mu}, \beta_{\mu}}(\mu^0_{x})$, $(\hat{\delta}^0_{x})^2 = Dropout_{p}((\delta_{x}^0)^2)$.
\STATE Sample $z^0 \sim \mathcal{N}(\hat{\mu}^0_{x},(\hat{\delta}_{x}^0)^2)$.
\STATE Transform $z^0$ into $z^T$ with $T$ IAF blocks in Equation~\ref{equ:IAF}.
\STATE Generate $x$ from $f_{\theta}(z^T)$.
\STATE Compute gradients $g_{\phi,\theta} \leftarrow \nabla_{\phi,\theta} \mathcal{L}_{ELBO}(x;\phi,\theta )$.
\STATE Update $\phi,\theta$, $\gamma_{\mu}$, $\beta_{\mu}$ according to $g_{\phi,\theta}$.
\STATE $\gamma_{\mu} = \frac{\gamma}{\sqrt{E_d[\gamma^2_{\mu_d}]}} \odot \gamma_{\mu}$
\ENDWHILE
\end{algorithmic}
\end{algorithm}

\noindent\textbf{$MPD_{p_{\mathcal{D}}(x)}[z^0]$ VS $MPD_{p_{\mathcal{D}}(x)}[z^T]$.} Here , we turn to further explore the possible connection between $MPD_{p_{\mathcal{D}}(x)}[z^0]$ and $MPD_{p_{\mathcal{D}}(x)}[z^T]$. Noting that $MPD_{p_{\mathcal{D}}(x)}[z^T]$ is intractable due to the interconnection between $h$, $z^0$ and each $\delta_t$, it is difficult to compare  $MPD_{p_{\mathcal{D}}(x)}[z^0]$ with $MPD_{p_{\mathcal{D}}(x)}[z^T]$ intuitively. However, we also find that when AutoregressiveNeuralNet in Equation~\ref{equ:IAF} ignores the context information $h$, i.e., the IAF chain is independent on input data, we can prove that $MPD_{p_{\mathcal{D}}(x)}[z^0]=MPD_{p_{\mathcal{D}}(x)}[z^T]$, which also demonstrates the strong relation between $z^0$ and $z^T$ in some degree.
\begin{proof}
First , we have the lemma 1~\cite{liese2006divergences}:
\begin{lemma}
Given two distributions $p_i(x)$ and $p_j(x)$ in space $\mathcal{X} (x \in \mathcal{X})$ and one differentiable and invertible transformation $h: \mathcal{X} \rightarrow \mathcal{Y}$ that converts $x$ into $y$, we have:
\begin{equation}
\small
\begin{split}
D_{KL} (p_i(x)||p_j(x)) = D_{KL}(p'_i(y)||p'_j(y)),
\end{split}
\end{equation} 
where $y=h(x)$ and $p'(y)$ denotes the distribution of $y$.
\end{lemma}

This lemma tells us that the KL-divergence between two distribution is invariant under one same differentiable and invertible transformation. 
Meanwhile, noting that IAF chains for different input data are all invertible and differentiable and same with each other when AutoregressiveNeuralNet ignores the context information $h$, we have:
\begin{equation}
\small
\begin{split}
D_{KL}(q(z^{t_1} | x_1) || q(z^{t_1}|x_2)) = D_{KL}(q(z^{t_2} | x_1) || q(z^{t_2} |x_2)),
\end{split}
\end{equation}
where $t_1,t_2 = 0,1,...,T$ and $x_1, x_2 \in \mathcal{X}$.
When we set $t_1=0$ and $t_2=T$, we can derive:
\begin{equation}
\small
\begin{split}
&MPD[z^0]=E_{p_{\mathcal{D}(x)}}[D_{SKL}[q_{\phi}(z^0|x_1)|q_{\phi}(z^0||x_2)]]\\
&=\frac{1}{2}E_{p_{\mathcal{D}(x)}}[D_{KL}[q_{\phi}(z^0|x_1)|q_{\phi}(z^0||x_2)] \\
&~~~~~~~~~~+ \frac{1}{2}E_{p_{\mathcal{D}(x)}}[D_{KL}[q_{\phi}(z^0|x_2)|q_{\phi}(z^0||x_1)]]\\
&=\frac{1}{2}E_{p_{\mathcal{D}(x)}}[D_{KL}[q_{\phi}(z^T|x_1)|q_{\phi}(z^T||x_2)] \\
&~~~~~~~~~~+\frac{1}{2}E_{p_{\mathcal{D}(x)}}[ D_{KL}[q_{\phi}(z^T|x_2)|q_{\phi}(z^T||x_1)]]\\
&= E_{p_{\mathcal{D}(x)}}[D_{SKL}[q_{\phi}(z^T|x_1)|q_{\phi}(z^T||x_2)]] =MPD[z^T].
\end{split}
\end{equation}

%
\end{proof}

\section{Detailed Experimental Setup}
Our experiments are designed with the guidance of~\cite{kim2018semi} and~\cite{he2018lagging}.  Here, we first introduce detailed experimental settings for experiments on real-world datasets,  and  the synthetic dataset, respectively. 
Finally, we will introduce more details of hyper-parameter selection for each baseline if needed.

\noindent\textbf{Text and Image Dataset.} In both Yahoo and Yelp dataset, we used the same train/val/test splits as provided by~\cite{he2018lagging}, which randomly downsample 100K/10K/10K sentences for training/validation/test, respectively. We utilized a single-layer LSTM  as both the encoder network and decoder network with 1024 hidden sizes. The size of word embeddings is 512.
Uniform distributions on $[-0.01,0.01]$ and $[-0.1,0.1]$ were applied for initializing the LSTM layers and embedding layers, respectively.
In addition, one Dropout of 0.5  was leveraged on both the input word embedding and the last dense for enhancing the performance. 
During the training process, the learning rate was initialized with 1.0 and decayed by 0.5 if the validation loss has not improved in the past 5 epochs.
Meanwhile, the training would stop early after 5 learning rate decays with the maximum number of epochs as 120.
As for OMNIGLTO dataset, we use the same train/val/test splits as provided by~\cite{kim2018semi}. When training, the inputs were binarized dynamically by sampling each pixel from one Bernouli distribution with the value in the original pixel as the parameter. Meanwhile,  the fixed binarization was utilized for validating and testing and the decoder uses binary likelihood. The 3-layer ResNet and 13-layer PiexlCNN used in our model is the same as these in~\cite{he2018lagging}. Adam optimizer was used for optimizing our model, which starts with a learning rate of 0.001 and decay it by 0.5 if the validation loss has not improved in the past 20 epochs. Meanwhile, the training would stop early after 5 learning rate decays.
In addition, following~\cite{kingma2016improved}, for all IAF based models, we used a 2-layer MADE~\cite{germain2015made} to implement each IAF block with hidden size 1920. Two IAF blocks were stacked as one IAF chain with ordering reversed between each other.

\noindent\textbf{Synthetic Dataset.} To generate the synthetic dataset with better visualization, we sample 2-dimensional latent variable $z$ from one mixture of Gaussian distributions that have 5 mixture components. Those Gaussian distributions are equipped with the mean parameters (0.0, 0.0), (-2.0, -2.0), (-2.0, 2.0), (2.0, -2.0) and (2.0, 2.0), respectively, and a unit variance. Then following~\cite{kim2018semi}, one LSTM layer with 100 hidden units and 100-dimensional input embeddings was used for generating synthetic text, where the hidden state was initialized   by the affine transformation of $z$. Then the output of the LSTM and $z$ were concatenated as the inputs of one MLP to map them into vocabulary space. The LSTM parameters were initialized with uniform distribution on [-1,1], and the parameters in MLP networks were initialized by uniform distribution on [-5,5]. We fixed the length of each text sample as 10 and the vocabulary size as 1000.
16000/2000/2000 examples were generated for training/validation/testing, respectively.
In the training procedure of different models, we utilized one single-layer LSTM as the encoder and decoder with 50 hidden units and 50 latent embeddings. The other experimental setup for synthetic experiments were the same as that for text datasets mentioned above.
In particular, to ensure meaningful representation, we selected the suitable parameters for BN-VAE and DU-VAE with minimal NLL when the corresponding MI is larger than 1 among hyper-parameters set $\gamma\in$\{0.5, 0.7, 1.0, 1.2\}, or $p\in$\{0.1, 0.3, 0.5, 0.7\}. Finally, we set $\gamma=1.0$ for BN-VAE and $\gamma=1.0$ and $p =0.5$ for DU-VAE in synthetic experiments.

 \begin{table}[!h]
\center
\small
\setlength{\tabcolsep}{1mm}{
\begin{tabular}{ccccc}\hline \hline
Model            & NLL          & KL  & MI  & AU  \\ \hline
BN-VAE with $\beta=0.1$      & 328.2      & 0.2 & 0.0 & 0.0  \\
BN-VAE with $\beta=0.3$    & 328.5        & 1.4 & 0.0 & 0.0  \\
BN-VAE with $\beta=0.5$    & 329.5       & 4.0 & 0.0 & 0.0  \\
BN-VAE with $\beta=0.7$    & 331.7       & 8.0 & 0.1 & 3.0 \\
BN-VAE with $\beta=1.0$    & 339.7       & 20.1 & 3.9 & 7.0 \\
\hline\hline
\end{tabular}}
\caption{The performance of likelihood estimation for BN-VAE with fixed non-zero $\beta$ on Yahoo.} 
\label{stab:nll}
\end{table}

\begin{table}[!h]
\center
\small
\setlength{\tabcolsep}{1mm}{
\begin{tabular}{cccccc}\hline\hline 
$\#$label       & 100  & 500  & 1k & 2k & 10k \\\hline
BN-VAE with $\beta=0.1$ &60.46&60.46&60.46&60.46&62.9 \\
BN-VAE with $\beta=0.3$ &60.41&60.46&60.46&60.46&62.41 \\
BN-VAE with $\beta=0.5$ &60.46&60.46&60.46&60.46&60.57 \\
BN-VAE with $\beta=0.7$ &60.46&60.46&60.49&60.53&62.78 \\
BN-VAE with $\beta=1.0$ &60.46&60.46&60.45&60.45&62.45 \\
 \hline\hline  
\end{tabular}}
\caption{The accuracy  of classification for BN-VAE with fixed  non-zero $\beta$ on Yelp.}
\label{stab:cla}
\end{table}

\noindent\textbf{More details for Hyper-parameter Selection.} For better Reproducibility, we show the specific ranges we considered to select the best hyper-parameters for each baseline and our models if needed. 
To be specific, for FB and IAF+FB, we varied the parameter $\lambda$ in  \{0.1, 0.15, 0.2, 0.25, 0.3\}. For $\delta$-VAE, we selected $\delta$ from \{0.1, 0.15, 0.2, 0.25\}. For BN-VAE and IAF+BN, the parameter $\gamma$ was selected from \{0.3, 0.4, 0.5, 0.6, 0.7\}. For MAE, we picked out parameter $\gamma$ and $\eta$ from \{0.5, 1.0, 2.0\}  $\times$ \{0.2, 0.5, 1.0\}. In addition, as for DU-VAE, we varied parameter $\gamma$ in \{ 0.4 ,0.5, 0.6, 0.7\} and $p$ in \{1.0, 0.9, 0.8, 0.7, 0.6\}. As for DU-IAF, we determined parameters $\gamma$ and $p$ from \{0.5, 0.6, 0.7, 0.8\} $\times$ \{0.9, 0.8, 0.7\} for text datasets and from  \{0.4, 0.5, 0.6, 0.7\} $\times$ \{0.9, 0.85, 0.8, 0.75\}  for image dataset due to the different performance of IAF based models on text and image datasets.
And all hyper-parameters are determined based on the NLL metric.


In addition, our code is based on Pytorch under Linux server,  and all experiments were conducted on  one NVIDIA Tesla v100 with 32GB RAM.
\section{Fixing Non-zero $\beta_{\mu}$ in BN}
Here, we aim to show one abnormal case to prove that keeping one positive lower bound of the KL term is not sufficient for the second term in Equation 10. To be specific, we explore the possibility of applying one BN on mean parameters $\mu$ of posteriors with fixed non-zero shift parameter $\beta_{\mu}$ and learnable scale parameter $\gamma_{\mu}$. We can find that this strategy can also ensure one positive lower bound of KL term, like BN-VAE. but, cannot avoid posterior collapse. 
To be specific, following the same experimental setting in Experimental Setup, we evaluated  the performance of likelihood estimation for this approach on Yahoo dataset and classification on the downsampled Yelp dataset by varying $\beta$ from $0.1$ to $1.0$. The results are summarized in Table~\ref{stab:nll} and~\ref{stab:cla}.
Comparing with Table S1 and S2, we find that the  BN strategy with fixed  non-zero $\beta$ fails to capture great performances on both classification and classification tasks, even keeps MI and AU metric vanishing to 0, which implies that the approximated posteriors of different input samples are same as each other. In other words, keeping one positive KL term is not sufficient for avoiding posterior collapse.

\section{Generation and Reconstruction}
\begin{figure*}[]
\centering
\subfigure[Samples generated from the prior distribution]{ 
\includegraphics[width=0.9\columnwidth]{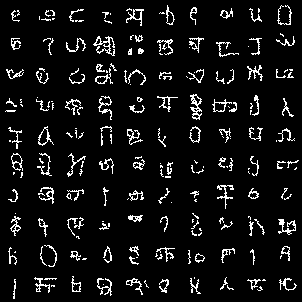} }
\subfigure[Interpolation between posterior samples]{ 
\includegraphics[width=0.99\columnwidth]{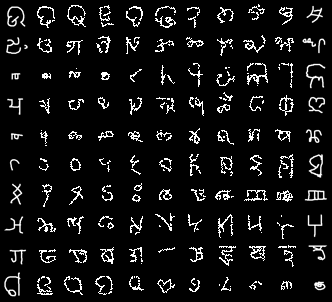} }
\caption{Generation and reconstruction on OMNIGLOT. The source and target images in Figure (b) are lied in the first and last columns, respectively.}
\label{fig:sam3}
\vspace{-4mm}
\end{figure*}

Here, we show some examples of generation and reconstruction for DU-VAE, which provide one illustrative perspective for the generative capacity. Specifically, we first generate examples guided by random samples selected  from the prior distribution.  Meanwhile, we also followed~\cite{bowman2015generating} and utilized linear interpolation between latent variables to evaluate the smoothness of the latent space. In particular, 
we displayed two samples from the test dataset and learn their latent representations $z_{x_1}$ and $z_{x_2}$ by the trained models. Then, we decoded greedily each linear interpolation point between those two with evenly divided intervals.  Table~\ref{tab:sam1} and~\ref{tab:sam2} show the results in the downsampled version of Yelp dataset, while the result for OMNIGLOT is summarized in Figure~\ref{fig:sam3}.
We can find that texts and images generated  by  DU-VAE are  grammatically plausible  or rich semantic. In addition, we  also note that  the examples decoded from  each interpolation point are semantically consistent  between source sample and target sample.

\begin{table*}[]
\center
\label{tab:sam1}
\begin{tabular}{ll}\hline\hline
This is a hidden gem.                                     & There was no excuse for the service , but no complaints. \\
They have the best pizza in town.                         & I got a new car and it was a complete waste of time.     \\
The staff was very friendly and helpful.                  & What a waste of time and money.                  \\
The food is delicious , and the service is very friendly. & We 'll definitely be back!                               \\
I love the food and the service here.                     & No thanks for the job.                                   \\
We will be back!                                          & The service is always good.                              \\
Food is good.                                             & The staff was very nice, and very helpful.    \\
This place is great.                                      & The lady who helped me was very rude.                    \\
This place is a great place to go!                        & It was a great experience.                               \\
No one ever came back to this place.                      & Thanks for the great service!      \\\hline\hline                     
\end{tabular}
\caption{Samples generated from the prior distribution on Yelp}
\end{table*}

\begin{table*}[]
\center
\label{tab:sam2}
\begin{tabular}{ll}\hline\hline
Source: I won't be back.                                    & Source: I was very disappointed with this place.        \\\hline
Target: I highly recommend this place.                      & Target: I love the atmosphere.                          \\\hline
I will not be back.                                         & I was very disappointed with this place.                \\
I would not recommend this place to anyone.                 & I was very disappointed with the service.               \\
I would recommend this place to anyone.                     & I am very disappointed with the service.                \\
I highly recommend this place.                              & I love the atmosphere.                                  \\
I highly recommend this place.                              & I love the atmosphere.                                  \\\hline\hline
Source: I got the chicken fajitas , which was a little dry. & Source: Service was great , and the food was excellent. \\\hline
Target: The food is always fresh and delicious.             & Target: My family and I love this place.                \\\hline
I got the chicken fajitas , it was a little dry.            & Service was great , and the food was great.             \\
I ordered the chicken sandwich , which was very good.       & Service was great and the food was great.               \\
I love the food here.                                       & My husband and I were here for a few years ago.         \\
The food is always fresh and delicious.                     & My husband and I love this place.                       \\
The food is always fresh and delicious.                     & My family and I love this place.              \\\hline\hline         
\end{tabular}
\caption{Interpolation between posterior samples on Yelp.}
\end{table*}

\vspace{-2mm}

\end{document}